\documentclass[10pt,twocolumn,letterpaper]{article}

\usepackage{wacv}
\usepackage{times}
\usepackage{epsfig}
\usepackage{graphicx}
\usepackage{amsmath}
\usepackage{amssymb}
\usepackage{booktabs}

\usepackage{multirow}
\usepackage{xcolor,colortbl}
\usepackage{subcaption}
\def\wacvPaperID{0013} 

\wacvapplicationstrack 

\wacvfinalcopy 

\ifwacvfinal
\usepackage[bookmarks=false]{hyperref}
\else
\usepackage[pagebackref=true,breaklinks=true,colorlinks,bookmarks=false]{hyperref}
\fi

\begin{document}

\title{A Computer Vision Method for Estimating Velocity from Jumps}

\author{Soumyadip Roy\(^1\), Chaitanya Roygaga\(^2\), Nathaniel Blanchard\(^1\), Aparna Bharati\(^2\)\\
\(^1\) Colorado State University, CO, USA \hspace{0.5cm} 
\(^2\) Lehigh University, PA, USA\\
}

\maketitle
\thispagestyle{empty}

\begin{abstract}
    Athletes routinely undergo fitness evaluations to evaluate their training progress. Typically, these evaluations require a trained professional who utilizes specialized equipment like force plates. For the assessment, athletes perform drop and squat jumps, and key variables are measured, e.g. velocity, flight time, and time to stabilization, to name a few. However, amateur athletes may not have access to professionals or equipment that can provide these assessments. Here, we investigate the feasibility of estimating key variables using video recordings. We focus on jump velocity as a starting point because it is highly correlated with other key variables and is important for determining posture and lower-limb capacity. We find that velocity can be estimated with a high degree of precision across a range of athletes, with an average R-value of 0.71 (SD = 0.06).
\end{abstract}
\section{Introduction}

Athletes benefit from feedback from professional evaluations to maximize performance and minimize injury risk. 
Often, these evaluations require athletes to perform key movements that facilitate evaluation, e.g. drop or squat jumps, with the use of specialized equipment like force plates. However, not all athletes have access to this equipment, nor to trained professionals to help them understand the variables measured by these movements and how they should adjust their training in response. Here, we establish that key variables from these jumps can be estimated using from videos of athletes performing both squat and drop jumps. 

We use force plate data as the ground truth for our variables. The accuracy of force plates has been well established. Garcia-Ramos \etal \cite{garcia2017reliability} show that the force plate variables like velocity and power are highly reliable when evaluating a countermovement jump for an athlete's performance. Velocity is an important variable when determining posture \cite{duarte2010revision} or to represent the maximum lower-limb capacity of an athlete \cite{samozino2014force}, which is helpful in creating individual and prioritized training profiles for athletes. 

\begin{figure}[htp!]
\begin{center}
   \includegraphics[width=0.45\textwidth]{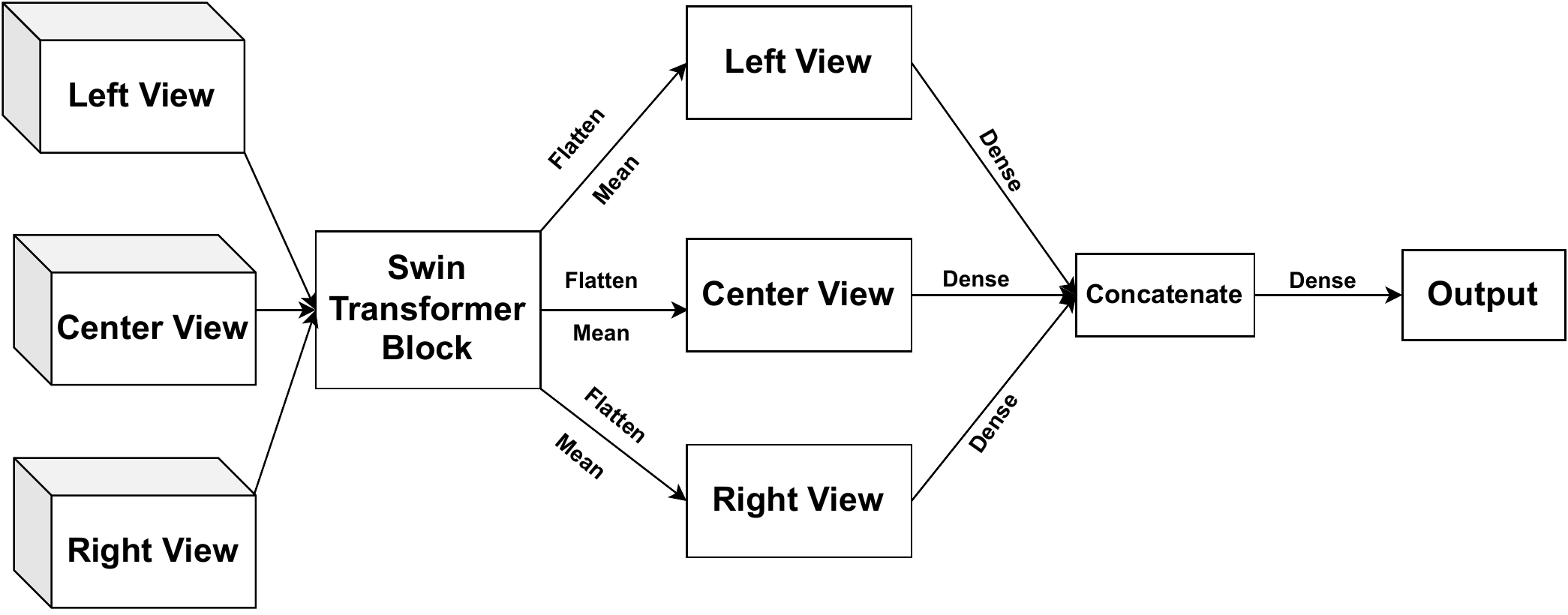}
\end{center}
   \caption{Overview of our pipeline. The Swin Transformer block takes as input the video frames and gives out flattened features further processed through two dense layers for an estimated velocity score.}
\label{fig:arch}
\end{figure}

The usefulness of force plates is limited to laboratory settings \cite{johnson2018predicting}, but it is substantially more feasible to record a video outside of such settings and utilize a computer vision model to estimate these key variables. However, for this to be realistic, the computer vision model has to be sufficiently reliable. Prior work has examined the feasibility of identifying when an athlete incorrectly performs a drop or squat jump \cite{roygaga2022ape}, but thus far no works have estimated the key variables themselves. 

Prior work has also focused on alternate solutions for athlete performance evaluation, useful in the absence of standard force plate data. Johnson \etal~\cite{johnson2018predicting} provide a method to analyze athlete ground reaction forces and moments (GRF/Ms) in real time using eight marker motion capture trajectories, allowing for a field-based study in identifying early injury risk; they \cite{johnson2019field} also use the motion capture data to estimate Knee Joint Movements (KJM). Goldacre \etal \cite{goldacre2021predicting} utilized preprocessed marker motion capture data and compared a Least Squares Estimator method for estimating GRF with a convolutional neural network (CNN) estimator and found the CNN-best method was more reliable and consistent. All of these works showcase the feasibility of estimating a key metric, GRF, but none have considered other essential metrics like velocity. Further, none of these works have attempted to estimate variables using raw video data. Here, we train a custom Swin Transformer architecture to estimate key metrics like velocity using the dataset from Blanchard \etal \cite{blanchard2019keep}.

We propose an automatic force plate variable estimation tool that uses athlete motion videos --- performing evaluative jumps such as countermovement and drop jumps --- 
as input, which can be run on any handy smart device such as a mobile phone. The tool is trained on the videos directly with minimal preprocessing steps, allowing faster inference during its execution. In the absence of a force plate, our method can provide a good approximation of the desired force plate variables, which can further be used to investigate the performance of an athlete.

In summary, this work has several novel contributions:
\begin{enumerate}
    \item This is the first work to showcase that key variables for assessing athlete performance can be estimated using video alone.
    \item We create a custom variant on the Swin Transformer architecture (see Figure \ref{fig:arch}) and achieve state-of-the-art performance across 86 athletes. 
    \item Our solution is generalized across two different kinds of jumps: the squat jump and the drop jump. We did not create separate models for each kind of jump.
\end{enumerate}

\section{Methods}

\subsection{Data}

Blanchard \etal \cite{blanchard2019keep} collected a dataset which consisted of 582 RGB videos of 89 athletes performing multiple countermovement and drop jumps~\cite{blanchard2019keep}. Each jump was recorded from three different angles: a left, right, and center perspective. Jumps were performed on force plate data, which recorded a multitude of variables associated with the jump (181 for countermovement jumps and 58 for drop jumps). Only 86 athletes' force plate data (545 videos) were available, so we only utilized this subset of the data. 

We divided the dataset into three separate folds. No participant from one fold was present in any other fold. The first two folds included 29 unique participants each, and the third fold included the remaining 28 participants. 
There were 180, 178, and 161 videos in each fold, respectively. 

\begin{figure}[htp!]
\begin{center}
   \includegraphics[width=1\linewidth,trim=4 4 4 4,clip] {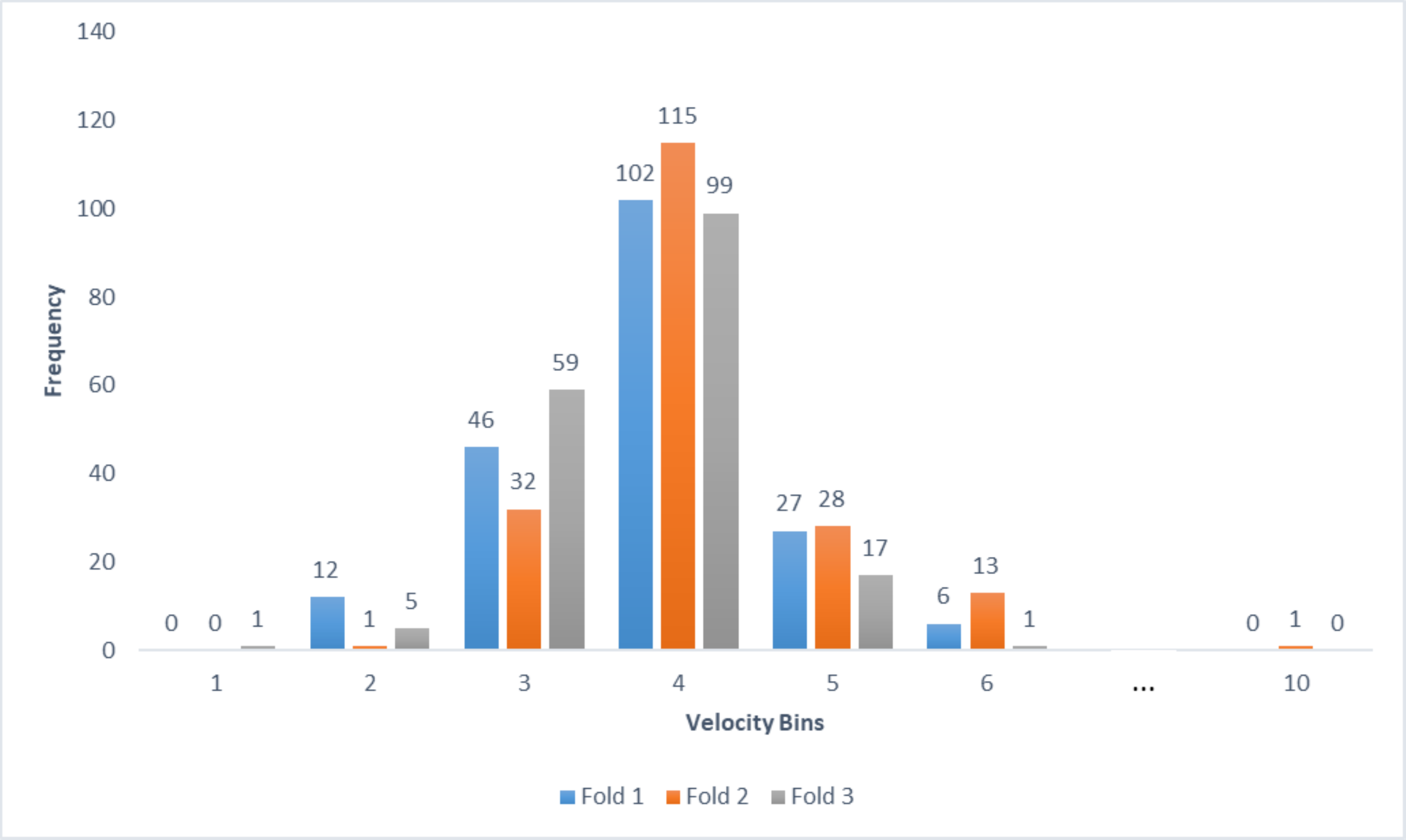}
\end{center}
   \caption{Distribution of velocity across data folds.}
\label{fig:Fold Graph}
\end{figure}

We focused on training a model to estimate concentric peak velocity. Figure~\ref{fig:Fold Graph} displays the distribution of concentric peak velocity (hereafter, velocity) across the three folds of our dataset. 
In our experiments, models used for evaluation were trained on each individual view, with an additional model trained on the combined input of the three views. 

\subsection{Model Architectures}

Estimating velocity, as well as other metrics from the force plate data, is a regression problem. We experimented with three different methods to evaluate the feasibility of estimating velocity.

\textbf{Baseline Models} We evaluated two different baseline methods for estimating the participant's velocity using video data in order to measure athletic performance without a force plate setup. First, we used a ResNet-50 architecture to extract features from all video frames. For our first baseline, we trained a support vector machine (SVM) using these features. For our second, we added a regression dense layer to the ResNet-50. 

In the first baseline method, we use the SVM regression method for velocity estimation, with the kernel set to 'rbf', degree to 3, coef0 to 0.0, tol to 0.0001, C to 1.0 and  epsilon to 0.01. The second baseline method uses a dense layer for final score estimation.

\textbf{Swin Transformer Architecture} We hypothesized a transformer architecture~\cite{dosovitskiy2020image} would be better suited for video data than our baseline methods. We modified the \textit{Swin Transformer}~\cite{liu2021swin} to extract features from video frames and utilized a dense layer to predict velocity. The first layer of the Swin Transformer divided the image into smaller patches with equal window sizes using the window multi-attention head (W-MSA), as shown in Equation~\ref{eq1}. The second layer repeated this, but shifted the starting position using shifted window multi-attention head (SW-MSA), as shown in Equation~\ref{eq2}. This change in how window patches are made across layers produced cross-connections that let the model identify the relevant image features. As illustrated in Equations~\ref{eq3} and~\ref{eq4}, self attention was conducted in the local window patches before the window patches were sent via a multi-layer perceptron (MLP). Finally, the adjacent patches were combined. The two successive layers were part of a Swin Transformer block; we used four such Swin Transformer blocks to generate the features with respect to each video frame.

\begin{equation}
\hat{z}^{l}= \mathrm{W}\\-\mathrm{MSA} \left( \mathrm{LN}\left( \mathrm{z}^{l-1}\right)\right) + \mathrm{z}^{l-1}
\label{eq1}
\end{equation}

\begin{equation}
\hat{z}^{l+1}= \mathrm{SW-MSA} \left( \mathrm{LN}\left( \mathrm{z}^{l}\right)\right) + \mathrm{z}^{l}
\label{eq2}
\end{equation}

\begin{equation}
z^{l}= \mathrm{MLP} \left( \mathrm{LN}\left( \mathrm{\hat{z}}^{l}\right)\right) + \mathrm{\hat{z}}^{l}
\label{eq3}
\end{equation}

\begin{equation}
z^{l+1}= \mathrm{MLP} \left( \mathrm{LN}\left( \mathrm{\hat{z}}^{l+1}\right)\right) + \mathrm{\hat{z}}^{l+1}
\label{eq4}
\end{equation}

where $\hat{z}^{l}$ are the output features of window-based multihead self-attention layers ($\mathrm{W}-\mathrm{MSA}$) and shifted window-based multihead self-attention layers ($\mathrm{SW}-\mathrm{MSA}$) for block $l$ respectively.

The architecture with the Swin Transformer backbone is as shown in Figure~\ref{fig:arch}. The inputs to the model are video frames from each view with respect to a single participant. Each of the video frames are processed through the Swin Transformer Block~\cite{liu2021swin}.

The mean of the frames from each view after flattening these features are calculated. Each view is passed through a dense layer before concatenating them all together. The combined features are then processed through another dense layer to obtain the final features. 

\subsection{Training and Evaluation Details}

Velocity scores followed a Gaussian distribution, with the majority of data ranging between 0.4 and 0.6, as seen in Figure \ref{fig:Fold Graph}. We exploited this distribution during training by using a balanced sampler to create mini batches, thus allowing the model to accurately learn the embedding space of jumps across all velocities.

All of our models were trained on a Titan V GPU with an Intel(R) Xeon(R) E5-1650 v4 processor and 16GB RAM. We used L1 loss for training and kept a fixed learning rate of 1e-3. The models were trained for 100 epochs using the Adam Optimizer. 

All models were trained and evaluated following a 3-fold cross validation paradigm. No participants present in the training data were present in the test data. 

We evaluated models using two metrics: Pearson's R and Mean Absolute Error (MAE). All reported results are the mean and standard deviation of each metric across all folds.

\section{Results}

\begin{table}[htbp]
\begin{center}
\begin{small}
\caption{Our baseline when compared to SVM and Dense Layers for estimation on the Video and Force plate data.}
\label{table:model_baseline_compare}
\begin{sc}
\renewcommand{\arraystretch}{1.3}
\begin{tabular}{| >{\raggedright\arraybackslash}p{2.7cm} | >{\raggedright\arraybackslash}p{2cm} | >{\raggedright\arraybackslash}p{2cm} |}
 \hline
 {\bfseries Method} & {\bfseries MAE ($\downarrow$)} & {\bfseries R  ($\uparrow$)}\\
 \hline
 {Baseline method with SVM} & {0.19 \textpm 0.017} & {0.12 \textpm 0.06}\\
 \hline
 {Baseline method with Dense Layer} & {0.28 \textpm 0.017 } & {0.16 \textpm 0.05 }\\
 \hline
 \hline
 \textbf{Our method (Combined view)} & {\bfseries 0.21 \textpm 0.015} & {\bfseries 0.71 \textpm 0.06}\\
 \hline
\end{tabular}
\end{sc}
\end{small}
\end{center}
\vspace{-0.2cm}
\end{table}

Table~\ref{table:model_baseline_compare} shows the performance of our baseline estimators and our transformer method. Although these methods appear similar when considering only MAE, our transformer-based model performs substantially better when evaluated by Pearson's R. In Figure 3, we showcase how our estimated velocity compares with the actual velocity for each fold --- our models slightly underestimated velocity, but they were extremely consistent with their estimations across folds. The low R values of the baseline methods indicate the estimations from those models were extremely inconsistent.

\begin{figure}[htp!]
     \centering
     \begin{subfigure}[b]{0.3\textwidth}
         \centering
         \includegraphics[width=\textwidth]{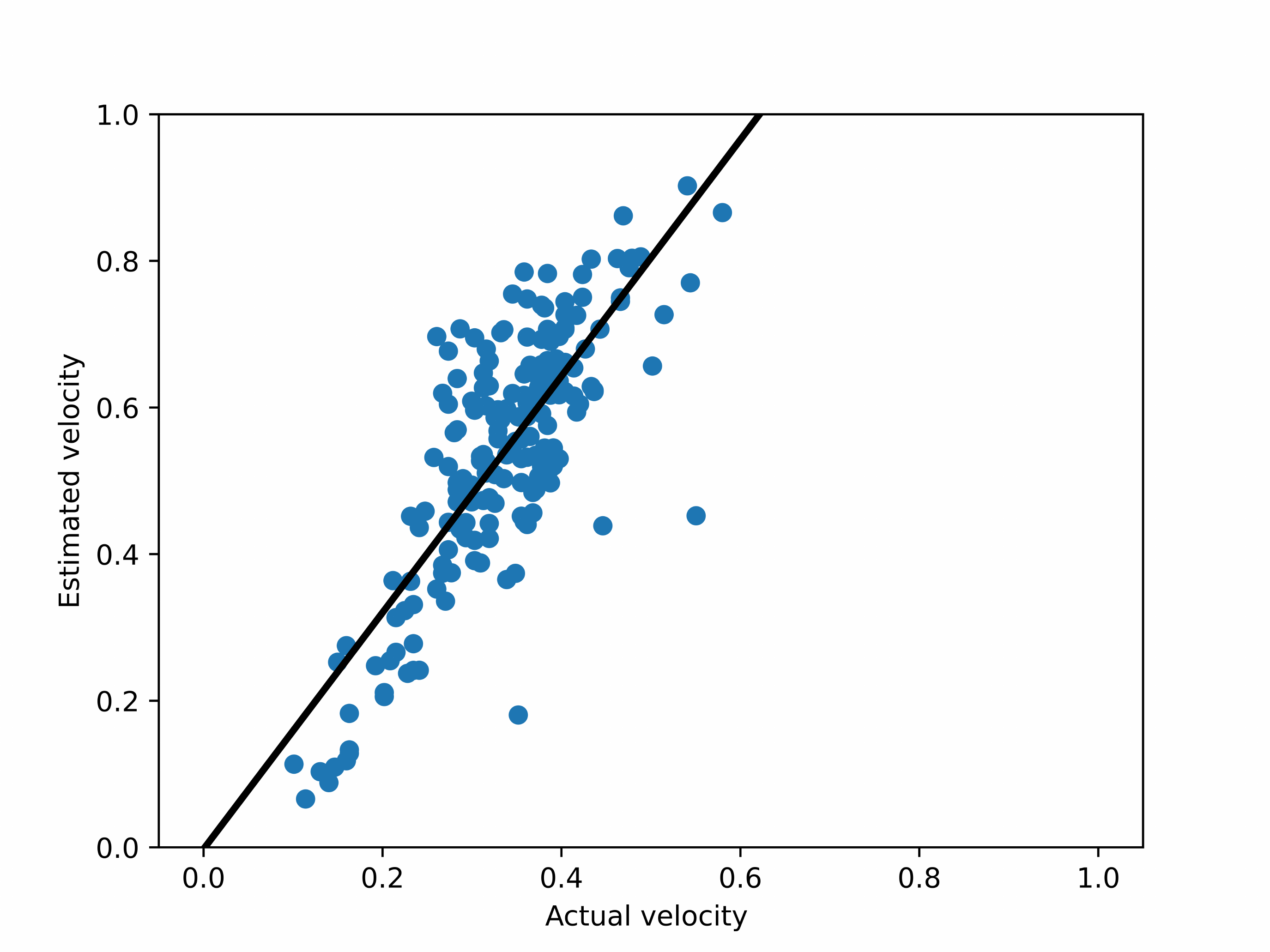}
         \caption{Fold 1}
         \label{fig:V1}
     \end{subfigure}
     \hfill
     \begin{subfigure}[b]{0.3\textwidth}
         \centering
         \includegraphics[width=\textwidth]{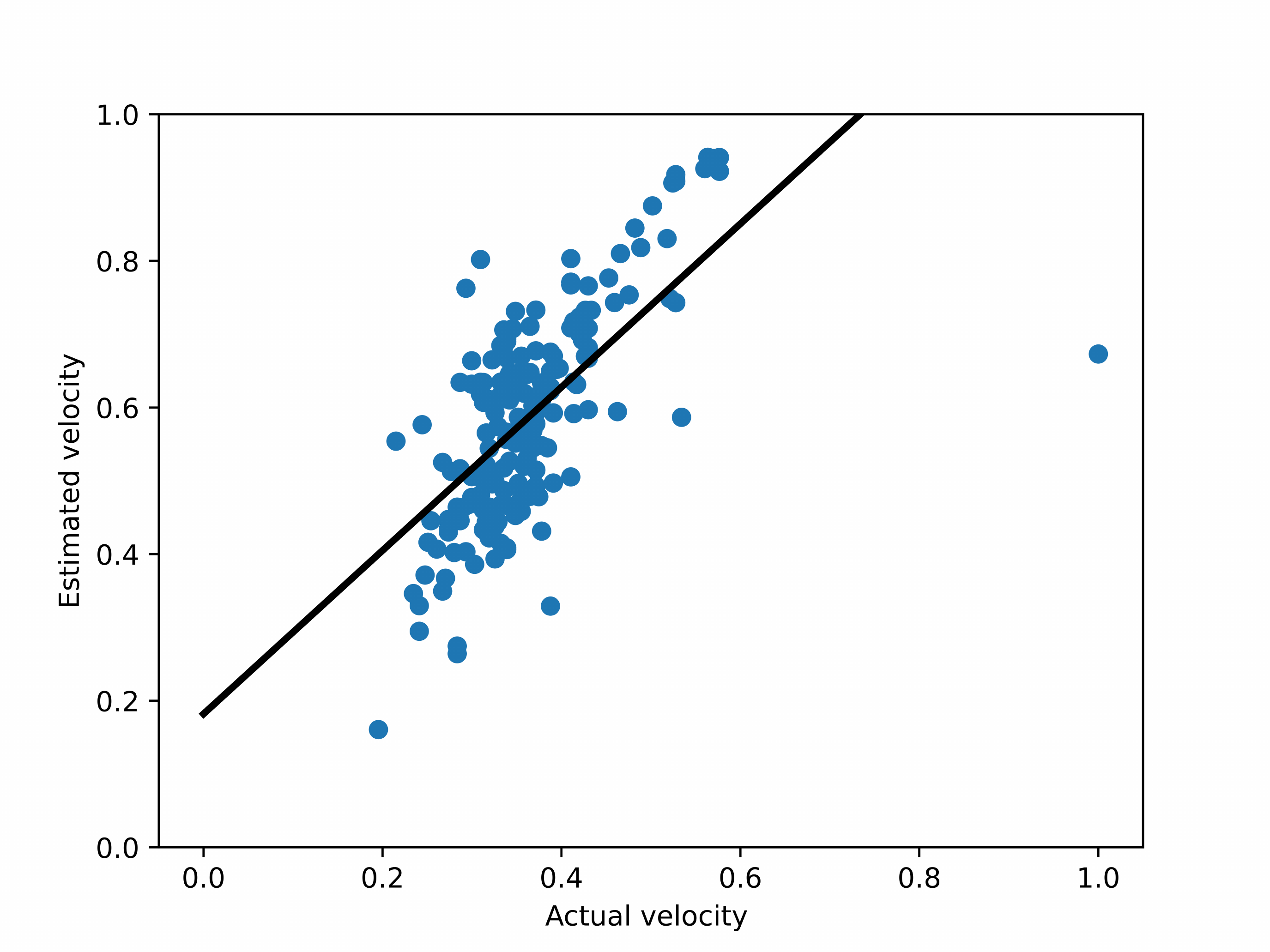}
         \caption{Fold 2}
         \label{fig:V2}
     \end{subfigure}
     \hfill
     \begin{subfigure}[b]{0.3\textwidth}
         \centering
         \includegraphics[width=\textwidth]{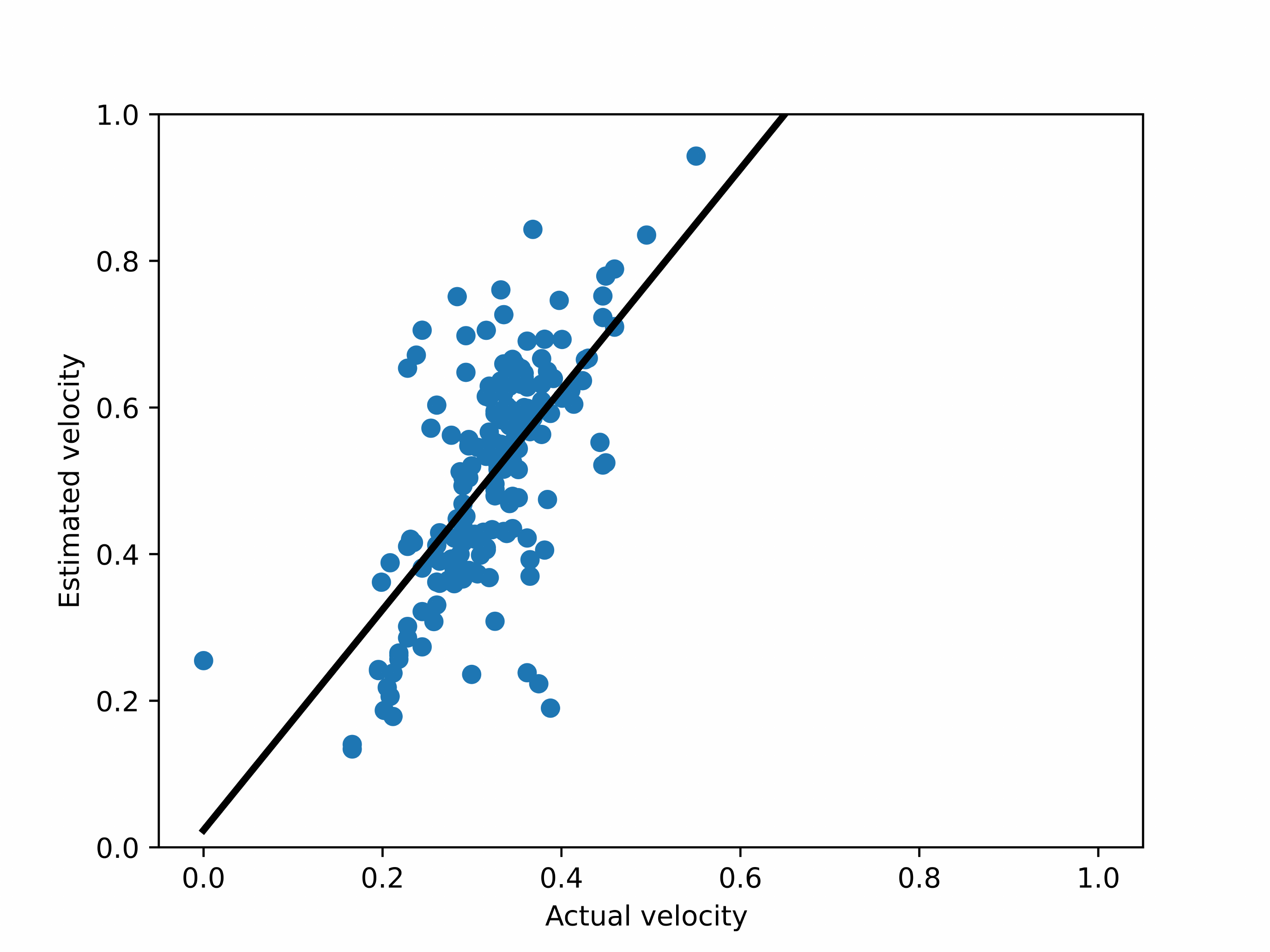}
         \caption{Fold 3}
         \label{fig:V3}
     \end{subfigure}
        \caption{Actual vs estimated velocity scores for the Transformer pipeline trained on a combined view input.}
        \label{fig:Fold Graph1}
\end{figure}

\subsection{Ablation}

We investigated how influential each component of the data was for our method's prediction. Table~\ref{table:model_view_compare} showcases the results of this experiment --- video from the center perspective was more important for accurately estimating velocity, exhibiting an R value 0.21 higher than either the left or right perspective (the left and right perspectives exhibited similar performances). This contrasts the findings of \cite{roygaga2022ape}, which found the left and right perspectives were more influential for estimating when a jump was performed incorrectly. This indicates that different perspectives may be essential to fully estimating all components of a jump. Further, data from all three perspectives culminated in the best model --- combining all perspectives increased the R value by 0.15 over the center view alone. 

\begin{table}[htbp] 
\begin{center}
\begin{small}
\caption{Experiments to determine the importance of combining videos captured from different views. The transformer pipeline trained on the combination of the three views improves the correlation score (R), which shows that the model is able to distinguish between a low and a high velocity jump when provided with video input from multiple angles. But the model trained on the Left view provides the least error in estimation, followed by the Right view, which shows that the model can still learn accurate estimations from videos from a single view but with a much lower capacity to distinguish between the higher and lower velocity jumps.}
\label{table:model_view_compare}
\begin{sc}
\renewcommand{\arraystretch}{1.3}
\begin{tabular}{| >{\raggedright\arraybackslash}p{2.7cm} | >{\raggedright\arraybackslash}p{2cm} | >{\raggedright\arraybackslash}p{2cm} |}
 \hline
 {\bfseries Video capture view} & {\bfseries MAE ($\downarrow$)} & {\bfseries R  ($\uparrow$)}\\
 \hline
 {Left} & {\bfseries 0.16 \textpm 0.016} & {0.35 \textpm 0.06}\\
 \hline
 {Right} & {0.20 \textpm 0.014} & {0.34 \textpm 0.04}\\
 \hline
 {Center} & {0.19 \textpm 0.014} & {0.56 \textpm 0.06}\\
 \hline
 {\textbf{Combined View}} & {0.21 \textpm 0.015} & {\bfseries 0.71 \textpm 0.06}\\
 \hline
\end{tabular}
\end{sc}
\end{small}
\end{center}
\vspace{-0.2cm}
\end{table}

\section{Conclusion and Future Work}

Velocity is a key evaluation metric for evaluating an athlete's performance. To our knowledge, this is the first work that estimates velocity from RGB video alone. Our novel solution is generalized across participants and two kinds of evaluative jumps: the squat jump and the drop jump. This work also provides insights into what kinds of data are important for estimating key metrics like velocity. In the future, we plan to expand our architecture to estimate a multitude of other essential variables for athlete performance following a multi-task regression paradigm. 

{\small
\bibliographystyle{ieee_fullname}
\bibliography{main}
}

\end{document}